# A Knowledge Graph based Solution for Entity Discovery and Linking in Open-domain Questions

Kai Lei[1], Bing Zhang[1], Yang Deng[1], Dongyu Zhang[2] and Ying Shen[1*]

[1] Institute of Big Data Technologies
Shenzhen Key Lab for Cloud Computing Technology & Applications
School of Electronic and Computer Engineering (SECE)
Peking University, SHENZHEN 518055 P.R.CHINA
[2] C.S. Depart. Harbin Institute of Technology
leik@pkusz.edu.cn, {zhang_bing,ydeng}@pku.edu.cn,
dongyu.z@hit.edu.cn, shenying@pkusz.edu.cn*

**Abstract.** Named entity discovery and linking is the fundamental and core component of question answering. In Question Entity Discovery and Linking (QEDL) problem, traditional methods are challenged because multiple entities in one short question are difficult to be discovered entirely and the incomplete information in short text makes entity linking hard to implement. To overcome these difficulties, we proposed a knowledge graph based solution for QEDL and developed a system consists of Question Entity Discovery (QED) module and Entity Linking (EL) module. The method of QED module is a tradeoff and ensemble of two methods. One is the method based on knowledge graph retrieval, which could extract more entities in questions and guarantee the recall rate, the other is the method based on Conditional Random Field (CRF), which improves the precision rate. The EL module is treated as a ranking problem and Learning to Rank (LTR) method with features such as semantic similarity, text similarity and entity popularity is utilized to extract and make full use of the information in short texts. On the official dataset of a shared QEDL evaluation task, our approach could obtain 64.44% F1 score of QED and 64.86% accuracy of EL, which ranks the 2nd place and indicates its practical use for QEDL problem.

## 1 Introduction

Question Answering (QA) is a popular research direction in Artificial Intelligence, aiming at building a system which can answer natural language questions automatically. Discovering entities in questions and linking them to the corresponding entries in the existing Knowledge Graph (KG) is the first step of QA because rich sources of facts from KG lays the foundation for answering the questions.

Specifically, Named Entity Discovery (or Recognition) (NED) is to discover and extract named entities from texts, which is critical technology of QA, information extraction, machine translation and many other applications. The concept of named entity was firstly proposed at Message Understanding Conference (MUC) [1], referring to the proper names (such as people names, place names, organization names) or other meaningful quantity phrases (such as time, date). In order to meet the needs of different applications, the meaning of named entities could be expanded. Entities such as product names, movie names etc. could also be included. Entity Linking (EL) [2] is to resolve named entities to corresponding entries in a structured KG. It can make full use of the semantic information of the rich text in knowledge graph, which has important significance in QA, information retrieval and knowledge graph construction.

To accelerate the development of related research, the China Conference on Knowledge Graph and Semantic Computing (CCKS) organized a shared evaluation task on Question Entity Discovery and Linking (QEDL) in 2017. QEDL is more difficult than traditional NED and EL tasks. Firstly, one short question may contains multiple entities, discovering all of them is a challenge. Secondly, it is difficult to obtain enough context information when linking entities to KG because questions are usually short texts. Moreover, only small amount of manual annotation training data is available sometimes, which requires the efficient method could converge quickly and obtain better results using less data.

To address the challenges mentioned above, we proposed a knowledge graph based solution for QEDL problem and developed a system consists of QED module and EL module. In QED module, the method based on KG retrieval was firstly employed, it could extract more entities in questions and guarantee the recall rate. Then the method based on Conditional Random Field (CRF) is utilized, which could improve the precision rate of entity discovery. Afterwards, two methods were merged together, which is a tradeoff of the precision and recall rate. Furthermore, the ensemble method could converge quickly to obtain ideal performance even if only small training corpus is available. EL module was treated as a ranking problem and Learning to Rank (LTR) method with features such as semantic similarity, entity popularity and text similarity is employed to make full use of the information in short texts.

The contributions of this paper can be summarized as follows:

- We proposed an ensemble method for high-density entity discovery in short questions, which could guarantee the recall rate without losses of the precision rate.
- In the ranking model of EL, rich features such as semantic similarity, entity popularity and text similarity are utilized to capture more information in short questions.
- We proposed a KG based solution and developed a system for QEDL problem, which is effective and converge quickly. The advantage is more obvious especially when the training set is relatively small.

The rest of this paper is structured as follows: Section 2 describes the related work. Section 3 introduces the details of the proposed methods. Experimental results and evaluations are presented in Section 4. Finally, we conclude this paper in section 5.

## 2   Related work

Lots of works have been involved in the research of NED and EL. The main technical methods of NED include rules and dictionary-based method, statistical method and the emerging method based on deep learning. Rules and dictionary-based method [3] is the earliest method used to NER task. But it has disadvantages such as long system construction period, time-consuming, poor portability and so on. Statistical method for NED uses machine learning models such as Hidden Markov Model (HMM) [4], Maximum Entropy (ME) [5], Support Vector Machine (SVM) [6] and CRF [7], etc. trained by manually annotated corpus. Thus the linguistic knowledge is not required. It can be completed in a short time and change less when transplanted into new domains. The method based on deep learning have been recently proposed, which include bidirectional Long Short-Term Memory with a CRF layer (BiLSTM-CRF) [8], BiLSTM and convolutional neural networks architecture (BiLSTM-CNN) [9] and other neural network models. Deep learning method doesn't need feature engineering, doesn't use any hand-crafted features or domain specific knowledge, thus it's portable. But it requires large amounts of manual annotation data and long training time. The evaluation of NED has been actively promoted the research. At present, the most influential evaluation meetings include Message Understanding Conference (MUC), Multilingual Entity Task Evaluation (MET), Automatic Content Extraction (ACE), Document Understanding Conference (DUC), etc.

Many of the entity linking systems use supervised machine learning methods, including LTR methods [10], graph-based methods [11] and model integration methods [12]. Vector Space Model (VSM) [13], as an unsupervised learning method, is also widely used in EL systems. In addition, many international meetings organized the evaluation of EL task, such as the "Link the Wiki" task in the EX meeting, the KBP task of the TAC meeting, the KBA tasks of the TREC meeting and the ERD'14 task at SIGIR. Although many researches have been carried out on the general domain, few studies focused on the question entity linking, which is more difficult because the information in questions is incomplete and has a lot of errors.

## 3   Methods

The QEDL task consists of two subtasks: QED and EL. Because of the small amount of training data, using joint learning method of the two subtasks is difficult to iterate until convergence and is prone to make mistakes. So we designed a pipeline system separating two subtasks with two independency modules.

In the QED module, we first proposed a concise and intuitive method, utilizing the n-gram model and former max matching algorithm to divide the sentence, and retrieve them in the knowledge graph. If a word or phrase is in the KG, then label it as candidate entity. Afterwards, we tried the method based on CRF and finally merged two methods together.

In the EL module, learning to rank method with rich features is used to rank the candidate entities and find the most matching one.

Fig.1 shows the overview of our system and details are described in this section.

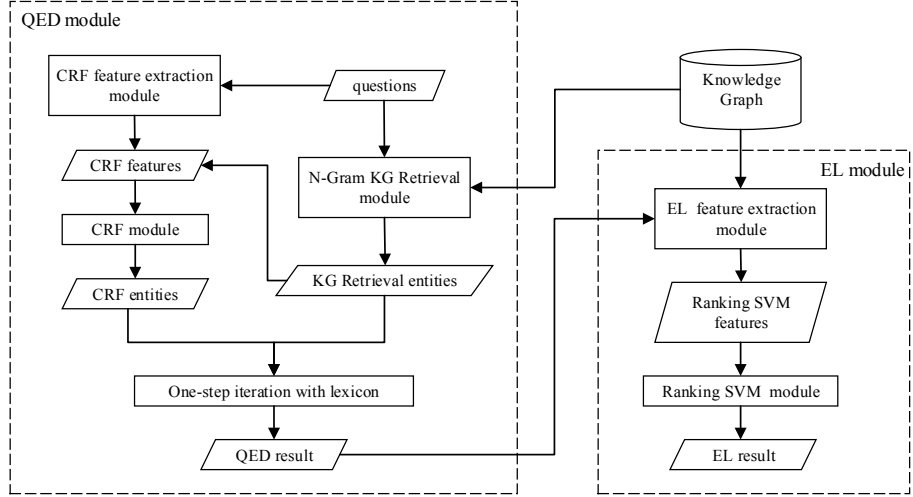

**Fig. 1.** Overview of our system

### 3.1 Question Entity Discovery

The QED module is an ensemble of two methods and details are as follows.

**Knowledge Graph Retrieval.** KG not only has a profound effect in the target field such as medical [14] and financial [15], but also contributes to open-domain QA. KG retrieval is widely used to generate candidate entities in question answering over knowledge graph [16]. The core idea is to search n-grams of words of the question to match the entity in the given knowledge graph and select a set of matching entities. We conduct this approach as following steps.

a. Generate all possible n-grams from the question, and tag parts of speech (POS);
b. Replace space and other meaningless symbols with a special mark "_";
c. Remove 1-grams that contains only one character;
d. Remove n-grams without any noun, verb, character or number;
e. Keep all the n-grams left which can match a certain entity in knowledge graph.

For example, a given question "孕妇吃方便面好吗?" generates a set of n-grams that match the entity in knowledge graph, like "孕妇/吃/方便/方便面/好/吗". After the procedure, three 1-grams, "吃/好/吗", are removed for containing only one character, while "方便" is discarded due to its POS tags with only adjective. Finally, we remain "孕妇/方便面" as question entities.

The KG retrieval method doesn't need feature extraction, training and testing and can obtain high recall rate, but the precision rate is relatively low.

**CRF.** CRF method regards QED as a sequence labeling problem. We utilized BIOES tagging rules in the sequence labeling system.

The CRF feature extraction module extracts features presented in Table 1.

**Table 1.** Feature list of QED module

| Feature name | Feature description |
| --- | --- |
| Character | A single character, and N-grams (N = 1,2,3,4) |
| Word boundary (WB) | The boundary of the word where the character is located |
| Part of speech (POS) | The part of speech of the word where the character is located |
| Stop words (SW) | The word where the character is located is stop words or not |
| Document frequency (DF) | The DF value of the word where the character is located |

Although features are simple, this method is effective, especially in terms of the precision rate.

**CRF based on KG retrieval.** In consideration of the high recall and low precision rate of KG retrieval method as well as the high precision and low recall rate of CRF method, we proposed a new ensemble method, CRF based on KG retrieval, which merge the two methods mentioned above together. More specifically, we tag the entities discovered in KG retrieval method with BIOES tagging rules and then take it as a feature of CRF.

The ensemble method is a tradeoff of the precision and recall rate and thus improved the system performance. On the one hand, it could improve the precision rate without much losses of recall rate comparing with the KG method, on the other hand, it could discover more entities with higher precision rate than traditional CRF method. The ensemble method can also obtain good result even though using less training data due to fast convergence.

**One-step Iteration with Lexicon.** As our system is a pipeline system of QED and EL, EL module uses the output of QED as input and the performance could be affected by QED results. We hope to improve the recall rate of QED to make sure more entities can be discovered and come into EL model. Thus one-step iteration using the result of KG retrieval and lexicon is added to our system. Those candidate entities discovered by KG retrieval method but ignored by CRF method would be matched to the lexicon. If the candidate entity is in the lexicon, then add it to the final discovery result. The lexicon we used is THUOCL [17] constructed by Tsinghua University.

Recall rate of QED is therefore improved by the iteration, which lays a solid foundation for the next step, EL module.

### 3.2 Entity Linking

Traditional EL module can be broken into two steps: candidate entity generation and candidate entity ranking. In this task, candidate entities can be generated using the provided API of CN-DBpedia, which is the knowledge graph constructed by Knowledge Works[1]. Therefore, our work mainly focused on candidate entity ranking. Ranking SVM is utilized to rank candidate entities and find the most matching one. To make full use of the information in short questions, rich features are employed and details are described below.

**Semantic Similarity.** The method we utilized to calculate semantic similarity between the question and candidate entity is Saliency-weighted semantic network proposed by [18]. The function for calculating semantic similarity is:

$$f_{ss} = \sum_{w \in q} IDF(w) \cdot \frac{sem(w,e) \cdot (k_1+1)}{sem(w,e) + k_1 \cdot (1-b+b \cdot \frac{|e|}{avge})} \quad (1)$$

$$sem(w,e) = \max_{w \in e} f_{sem}(w,w') \quad (2)$$

Here, $q$ is the question, $w$ is the term in $q$, $e$ is the candidate entity and $avge$ is the average length of candidate entities. $IDF(w)$ calculated from large amount of unlabeled Wiki corpus is used to weight the words in questions based on the idea that common terms (like determiners) do not contribute as much to the meaning of a text as less frequent words do. In formula (2), $sem(w,e)$ is the semantic similarity of term $w$ with respect to the candidate entity $e$. The function $f_{sem}$ returns the semantic similarity between two terms. As terms are represented as vectors using word embeddings trained by Wiki corpus, $f_{sem}$ could be calculated by the distance between two vectors, which reflects semantic similarity information. Cosine similarity is used to calculate distance between vectors in our system. The parameters k1 and b have a smoothing effect and we default set k1 = 1.5 and b = 0.75.

Formula (1) looks similar to the famous BM25 formula [19], but original BM25 formula only captures the lexical similarity between two texts, while we implement the formula with TF-IDF weighting scheme and word embeddings to measure both lexical and semantic similarity between two texts.

**Text Similarity.** Term Frequency-Inverse Document Frequency (TF-IDF) model [20], Latent Semantic Indexing (LSI) model [21] and Latent Dirichlet Allocation (LDA) model [22] are effective and frequently used methods for text similarity calculation.

TF-IDF model convert text into fixed-length vector space and spatial similarity is used to approximate text similarity. Words in the text are weighted by the number of occurrences in the text and the importance to the text. LSI uses Singular value decomposition (SVD) technique to word-document matrix to reduce the dimension of TF-IDF model. LDA is the topic model, the word vectors of texts after remove stop words are mapped to the topic distribution and cosine similarity is calculated to represent the text similarity. Gensim[2] is used to build TF-IDF, LSI and LDA model.

---

[1] http://kw.fudan.edu.cn/
[2] http://radimrehurek.com/gensim/index.html

The three methods above are exploited to calculate text similarity between questions and the name of candidate entities and the values were put together as a feature set of learning to rank model. In addition to the text similarity between questions and entity name (TS_QEN for short), text similarity between questions and the attributes of candidate entities obtained by API (TS_QEA for short) is also calculated.

**Entity Popularity.** The popularity of an entity indicates the possibility of the entity being mentioned in a question. We use the number of results returned by search engine when searching the entity to represent entity popularity. The popularity feature is defined as follows:

$$P(e) = logN \qquad (3)$$

Given an entity e, N is the hit number returned by Baidu. For example, the entity mention "方便面" corresponds to two candidate entities in CN-DBpedia, "方便面（快餐类面制食品）" and "方便面（中国大陆歌手肖飞演唱歌曲）". When we search them respectively in Baidu, the former retrieves about 1,370 relevant results while the latter retrieves about 447 relevant results. The popularity of candidate entities proved to be a distinguishable feature to EL task.

## 4 Experiments and evaluation

Experiments and evaluation have been carried out based on the training set which contains about 1400 manually annotated questions and the test set contains about 800 questions without labels published by CCKS2017 QEDL task. The knowledge graph this task uses is CN-DBpedia, which contains hundreds of millions of entities and could be accessed through API. The evaluation results are as follows.

### 4.1 Question Entity Discovery Results

QED is treated as a sequence labeling problem in our system and different methods with different features described in Section 3.1 are exploited. To evaluate the results, Precision rate, Recall rate and F1 Score are used as evaluation indicators in QED module. The results of the experiment are shown in Table 2.

Table 2. Performance of QED module

| Methods | Features | Precision (%) | Recall (%) | F1 (%) |
|---|---|---|---|---|
| KG Retrieval | / | 28.63 | 72.60 | 41.06 |
| CRF | character | 44.66 | 43.28 | 43.96 |
|  | character+WB | 46.95 | 50.11 | 48.48 |
|  | character+WB+POS | 46.46 | 53.84 | 49.88 |
|  | character+WB+POS+SW | 47.42 | 53.94 | 50.47 |
|  | character+WB+POS+SW+DF | 47.88 | 54.26 | 50.87 |

| CRF based on KG retrieval | character+WB+POS+SW+DF+ KG information | **55.90** | 67.16 | 61.02 |
| One-step Iteration with Lexicon | character+WB+POS+SW+DF+ KG information | 55.36 | **77.08** | **64.44** |

As is presented in Table 2, KG retrieval method could obtain high recall rate, but the precision rate is low. Traditional CRF method with features such as character, word boundary, part of speech, stop words and document frequency of terms obtained higher precision rate and F1 score compare with KG Retrieval method, but the recall rate is relatively low. The CRF based on KG retrieval method is really effective, it has positive effect on both precision and recall rate and increases 10.15 percent point of the F1 score on the foundation of traditional CRF method. At last, although one-step iteration with lexicon couldn't increase the precision rate, it has greatly improved the recall rate and thus improved the F1 score, which also lays a solid foundation for the next step, EL module.

In addition to the quality of entity discovery, convergence speed of methods should also be concerned about, especially when only a small amount of labeled training data is available. To evaluate the convergence speed, we developed an experiment utilizing different size of training sets and different methods. The methods being evaluated in this experiment include traditional CRF, CRF based on KG Retrieval proposed in this paper and BiLSTM-CRF, the emerging and outstanding deep learning method for entity discovery. The result is shown in Fig.2 and Fig.3.

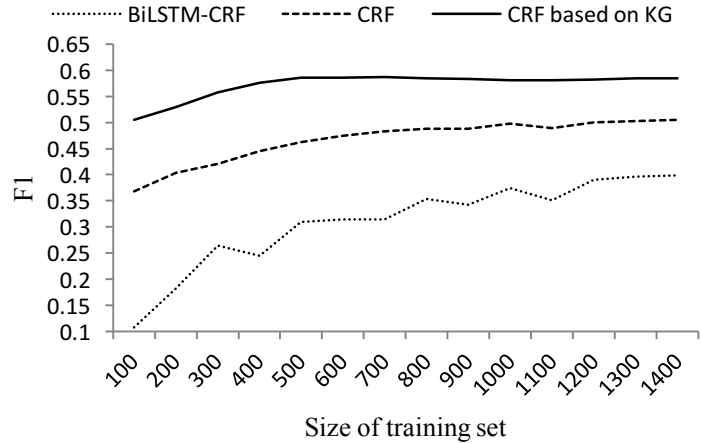

**Fig. 2.** Comparisons of convergence speed in different methods

Fig.2 illustrates that traditional CRF method could converge when the size of training set is about 1000 while the method we proposed, CRF based on KG Retrieval, could converge utilizing only about 600 training data, which is very efficient. As for BiLSTM-CRF method, it is hard to converge on small training set thus the performance on this task is unsatisfactory, it requires much larger size of manually annotated training data.

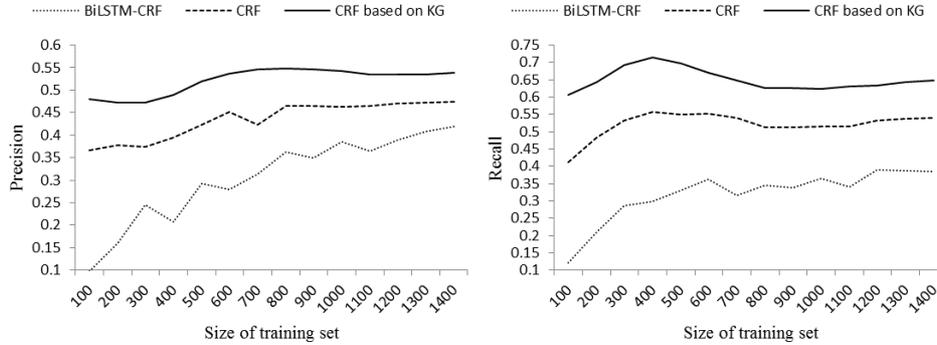

**Fig. 3.** Tendency of Precision and Recall rate with the change of training set

Fig.3 illustrates the tendency of precision and recall rate with the change of training set in different methods. CRF based on KG Retrieval method still converges most quickly. Meanwhile, the tendency also supports the perspective that CRF based on KG Retrieval method, as an ensemble method, is a tradeoff of the precision and recall rate and thus improved the overall performance of QED module.

### 4.2 Entity Linking Results

In the EL module, ranking SVM with features described in Section 3.2 is employed to rank the candidate entities. In order to evaluate the performance of EL without being disturbed by the result of QED, we only evaluate the EL performance of those correctly recognized entities. Obviously, the Precision=Recall=F1=Accuracy of EL under the premise that the correct entity mention is given. Table 3 shows the experimental results of EL module.

**Table 3.** Performance of EL module

| Features | Accuracy (%) |
| --- | --- |
| Semantic Similarity | 60.71 |
| Semantic Similarity + TS_QEN | 61.55 |
| Semantic Similarity + TS_QEN + TS_QEA | 62.37 |
| Semantic Similarity + TS_QEN + TS_QEA + Entity Popularity | **64.86** |

From Table 3 we can see, performance of EL module is improved with the increase of the feature sets, each of them can capture different aspects of information in questions and candidate entities and thus contribute to the result.

In addition to evaluating the EL module, it is interesting to see how different sets of features affect the performance. To analyze the importance of one feature set, we leave out it and use the rest of features to calculate the result. The results of leaving out different feature sets are shown in table 4, sorted by accuracy.

**Table 4.** Effect of each feature set

| Omitted feature set | Accuracy (%) |
|---|---|
| Semantic Similarity | 60.44 |
| TS_QEN | 61.13 |
| TS_QEA | 61.55 |
| Entity Popularity | 62.79 |

Table 4 shows that leaving out the semantic similarity feature has the most dramatic effect on performance. So the semantic similarity feature has the most significant contribution to the ranking model, next is TS_QEN and TS_QEA feature sets, the entity popularity feature contributes least.

### 4.3 Overall Results

At last, we evaluate the overall performance of the entire QEDL system. The evaluation results are shown in Table 5.

**Table 5.** Overall performance of QEDL

| NED (%) | | | EL (%) | Overall (%) | | |
|---|---|---|---|---|---|---|
| Precision | Recall | F1 | Accuracy | Precision | Recall | F1 |
| 55.36 | 77.08 | 64.44 | 64.86 | 38.96 | 54.05 | 45.28 |

The overall performance of our method ranks the 2nd place in CCKS2017 QEDL task, indicating its practical use for QEDL problem.

## 5 Conclusion

This paper introduces a knowledge graph based solution of QEDL problem consists of QED and EL module. In the QED module, CRF based on knowledge graph retrieval with one-step iteration method is utilized, which could discover multiple entities in questions without losses of precision and converge quickly with small size of training set. EL module is treated as a ranking problem and ranking SVM with semantic similarity, text similarity and popularity features is employed to make full use of the information in short texts. The results of evaluation show that our approach could converge faster than BiLSTM-CRF method in QED and obtain higher F1 score up to 64.44% while the accuracy of EL is 64.44%, which ranks the 2nd place in QEDL evaluation task. According to the result, our solution is valuable, especially when labeled data set is not very adequate. In the future we want to extend our system to more NED and EL problems not only in questions but also in other short texts.

## References

1. Chinchor, N. MUC7 Named Entity Task Definition. Message Understanding Conference (1997).

2. Han, X., Sun, L.: A generative entity-mention model for linking entities with knowledge base. In: The Meeting of the Association for Computational Linguistics: Human Language Technologies, Proceedings of the Conference, 19-24 June, 2011, Portland, Oregon, Usa, pp.945-954 (2011).
3. Humphreys, R. G., Azzam, S., Huyck, C., Mitchell, B., Cunningham, H., Wilks, Y. Description of the LaSIE-II System as Used for MUC7, pp.127-140 (1998)..
4. Fu, G., Luke, K.K.: Chinese named entity recognition using lexicalized HMMs. In: ACM SIGKDD Explorations Newsletter 7(1), pp.19-25 (2005).
5. Hai, L. C., & Ng, H. T. Named entity recognition: a maximum entropy approach using global information. In: International Conference on Computational Linguistics pp.1-7 (2002).
6. Li, L., Mao, T., Huang, D., Yang, Y.: Hybrid models for Chinese named entity recognition. In: Proceedings of SIGHAN Workshop, pp. 72-78 (2006)
7. Chen, A., Peng, F., Shan, R., Sun, G.: Chinese named entity recognition with conditional probabilistic models. pp. 173-176 (2006).
8. Huang, Z., Xu, W., Yu, K.: Bidirectional LSTM-CRF models for sequence tagging. In: Computer Science, (2015).
9. Chiu, J. P. C., Nichols, E.: Named entity recognition with bidirectional LSTM-CNNs. In: Computer Science, (2015).
10. Zheng, Z., Li, F., Huang, M., Zhu, X.: Learning to link entities with knowledge base. In: The North American Chapter of the Association for Computational Linguistics: Human Language Technologies, pp. 483-491 (2010)
11. Hoffart, J., Yosef, M. A., Bordino, I., Fürstenau, H., Pinkal, M., Spaniol, M.: Robust disambiguation of named entities in text. In: EMNLP, pp.782-792 (2011).
12. Mihalcea, R., & Csomai, A.: Wikify!:linking documents to encyclopedic knowledge. In: CIKM, pp.233-242 (2007).
13. Cucerzan, S.: Large-scale named entity disambiguation based on wikipedia data. In: EMNLP-CoNLL, pp.708-716 (2007).
14. Jayaraman, S., Tao, L., Gai, K., & Jiang, N.: Drug Side Effects Data Representation and Full Spectrum Inferencing Using Knowledge Graphs in Intelligent Telehealth. In: IEEE, International Conference on Cyber Security and Cloud Computing, pp.289-294 (2016).
15. Elnagdy, S. A., Qiu, M., & Gai, K.: Cyber Incident Classifications Using Ontology-Based Knowledge Representation for Cybersecurity Insurance in Financial Industry. In: IEEE, International Conference on Cyber Security and Cloud Computing, pp.301-306 (2016).
16. Golub D, He X.: Character-level question answering with attention. In: Proceedings of EMNLP, pp. 1598-1607 (2016)
17. Han, S., Zhang, Y., Ma, Y., Tu, C., Guo, Z., Liu, Z., Sun, M.: THUOCL: Tsinghua open Chinese lexicon. (2016).
18. Kenter, T., Rijke, M. D.: Short text similarity with word embeddings. In: CIKM, pp.1411-1420 (2015).
19. Robertson, S., Zaragoza, H.: The probabilistic relevance framework: bm25 and beyond. In: Foundations & Trends® in Information Retrieval, 3(4), pp.333-389 (2009).
20. Xu, W.: A Chinese keyword extraction algorithm based on TFIDF method. In: Information Studies Theory & Application (2008).
21. Mirzal, A.: Similarity-based matrix completion algorithm for latent semantic indexing. In: IEEE ICCSCE, pp.79-84, (2014).
22. Celikyilmaz, A., Hakkani-Tur, D., Tur, G.: LDA based similarity modeling for question answering. In: Proceedings of the NAACL HLT Workshop on Semantic Search, pp.1-9 (2010).